%% file: template.tex
\documentclass{vgtc}                          

\graphicspath{{figures/}{pictures/}{images/}{./}} 

\usepackage{times}                     

\usepackage{tabu}                      
\usepackage{booktabs}                  
\usepackage{lipsum}                    
\usepackage{mwe}                       

\usepackage{mathptmx}                  

\usepackage{verbatim}
\usepackage{graphicx}
\usepackage{bibentry}
\usepackage{amsmath}
\usepackage{multirow}
\usepackage{booktabs}
\usepackage{tabularx}
\usepackage{makecell}
\usepackage{bbding}
\usepackage{pifont}
\usepackage{wasysym}
\usepackage{amssymb}
\usepackage{svg}
\usepackage{rotating}
\usepackage[export]{adjustbox}
\usepackage{algorithm}
\usepackage{algpseudocode}

\newcommand{\ie}{\textit{i.e.}}

\newcommand{\etal}{\textit{et al.}}

\definecolor{watermelon_green}{rgb}{0, 0.42, 0.30}

\newcommand{\HS}[1]{\textcolor{black}{#1}}

\onlineid{0}

\vgtccategory{Research}

\vgtcinsertpkg

\title{MC-INR: Efficient Encoding 
of Multivariate Scientific Simulation Data 
using Meta-Learning and Clustered Implicit Neural Representations}

\author{
Hyunsoo Son \thanks{e-mail: ehsoon@korea.ac.kr}\\ %
    \scriptsize Korea University %
\and Jeonghyun Noh \thanks{e-mail: wjdgus0967@korea.ac.kr} \\
    \scriptsize Korea University %
\and Suemin Jeon \thanks{e-mail: orangeblush@korea.ac.kr} \\
    \scriptsize Korea University %
\and Chaoli Wang \thanks{e-mail: chaoli.wang@nd.edu} \\
    \scriptsize University of Notre Dame %
\and Won-Ki Jeong \thanks{e-mail: wkjeong@korea.ac.kr} \\
    \scriptsize Korea University
}

\abstract{

    Implicit Neural Representations (INRs) are widely used to encode data as continuous functions, enabling the visualization of large-scale multivariate scientific simulation data with reduced memory usage.
    However, existing INR-based methods face three main limitations: (1) inflexible representation of complex structures, (2) primarily focusing on single-variable data, and (3) dependence on structured grids.
    Thus, their performance degrades when applied to complex real-world datasets.
    To address these limitations, we propose a novel neural network-based framework, MC-INR, which handles multivariate data on unstructured grids.
    It combines meta-learning and clustering to enable flexible encoding of complex structures.
    To further improve performance, we introduce a residual-based dynamic re-clustering mechanism that adaptively partitions clusters based on local error.
    We also propose a branched layer to leverage multivariate 
    data
    through independent branches simultaneously.
    Experimental results demonstrate that MC-INR outperforms existing methods on scientific data encoding tasks.
} 


\begin{document}


\firstsection{Introduction}
\maketitle

\input{section/1_intro}

\section{Related Work}

\input{section/2_related_work}

\section{Proposed Method: MC-INR}

\input{section/3_method}

\section{Experiments}

\input{section/4_exp}

\section{Conclusion}

\input{section/5_conclusion}

\section*{Acknowledgments}
This work was partially supported by the National Research Foundation of Korea (NRF) grant funded by the Korea government (MSIT) (RS-2024-00349697), the National Research Council of Science \& Technology (NST) grant by MSIT (No.~GTL24031-000), the ICT Creative Consilience program of the Institute for Information \& communications Technology Planning \& Evaluation (IITP) funded by MSIT (IITP-2025-RS-2020-II201819), and a Korea University Grant.

\bibliographystyle{template/abbrv-doi-hyperref-narrow}

\bibliography{template}
\end{document}

%% file: section/1_intro.tex
\label{sec:intro}

Scientific simulations numerically 
model
spatiotemporal variations to capture complex physical phenomena.
%
The resulting data typically consists of multiple variables or features, such as temperature, pressure, velocity, etc., that vary continuously over spatiotemporal coordinates. 
%
Visualizing such multivariate scientific data is essential to recognize patterns and outliers at particular spatial or temporal locations, which has a long-standing history of research~\cite{Lachmann2012}. 
%
%
Traditionally, multivariate scientific data have been represented using grids or polygonal meshes~\cite{muigg2007scalable,shen2006visualization,wang2018visualization,yu2010situ}. 
However, recent neural network-based methods, such as implicit neural representations (INRs)~\cite{han2022coordnet,jiao2024ffeinr,lu2024fcnr,lu2021compressive,tang2023ecnr,tang2024stsr}, have demonstrated remarkable performance, particularly in data compression and super-resolution, effectively reducing data size.
%
%
Notably, INRs are well-suited for scientific visualization, as they approximate values at given spatiotemporal coordinates and represent them as continuous functions.


However, conventional INRs face several limitations in representing multivariate scientific simulation data. 
%
First, most existing INR methods rely on a single network to represent the entire domain, which restricts their ability to capture complex spatial structures and shapes.
%
%
%
Second, they are primarily designed to handle single-variable data, limiting their ability to represent multivariate scientific simulations effectively.
In real-world scientific simulation data, multiple physical variables interact within the same spatial domain, making it challenging for traditional representations to capture such complexity, often resulting in poor performance.
%
%
%
Lastly, INRs for representing time-varying scientific data are generally biased toward structured grids, making it challenging to represent complex geometric structures in unstructured grids. 
As a result, they often fail to capture intricate spatial structures accurately.


To address these limitations, we propose \textbf{MC-INR} (\ie, meta-learning and clustered INR), a novel neural network-based representation that extends CoordNet~\cite{han2022coordnet} for efficient encoding of multivariate data defined on unstructured grids.
%
%
Our contributions are as follows: First, we combine meta-learning~\cite{yang2025meta} and spatial clustering to enable efficient encoding of local patterns and flexible adaptation to complex structures.
MC-INR partitions the data into spatial clusters and learns cluster-specific meta-knowledge.
%
%
We also introduce a residual-based dynamic re-clustering mechanism that adaptively partitions clusters to more accurately capture structural variations with reduced error.
%
%
Second, we propose a branched network architecture inspired by multi-task learning, in which each variable is assigned a dedicated branch to capture fine-grained structures, thereby improving the representation accuracy of multivariate data.
Notably, the proposed network and spatial domain decomposition make our method particularly well-suited for representing multivariate data on unstructured grids.
%
We demonstrate the effectiveness of MC-INR both quantitatively and qualitatively using our in-house small modular reactor simulation datasets. 


%% file: section/2_related_work.tex
\subsection{Meta Learning}
\label{sec:rel_meta}
Meta-learning~\cite{finn2017model} is a training framework designed to enable models to quickly adapt to new tasks and complex structures.
In particular, it has been widely applied in INRs to reduce training time and improve generalization performance.
Sitzmann \etal~\cite{sitzmann2020metasdf} proposed a gradient-based meta-learning approach, and Tancik \etal~\cite{tancik2021learned} introduced a method that adaptively targets classes of signals, leveraging the benefits of meta-learning.
Yang \etal~\cite{yang2025meta} proposed Meta-INR, which improves the generalization and representation of time-varying data by interval-selecting points for training the meta-learner.
Inspired by these strategies, we adopt meta-learning to enable efficient training and represent complex structures.

\subsection{Clustering}
\label{sec:rel_clu}
Several scientific simulation datasets are large in scale, making it difficult to fully encode and utilize their complex features due to high computational memory requirements.
To mitigate this limitation, various methods employed tree-based~\cite{martel2021acorn,yang2023sci} or Laplacian pyramid-based~\cite{saragadam2022miner,tang2023ecnr} approaches to partition the data into smaller, more manageable clusters.
Liu \etal~\cite{liu2024uginr} used k-means clustering to apply unstructured grids, addressing cases that previous methods find challenging to handle.
Inspired by this, we employ the k-means clustering to treat unstructured grids; furthermore, each cluster is processed on a separate GPU to improve training speed and computational efficiency.

\begin{figure*}[t]
    \centering
    \includegraphics[width=0.95\textwidth]{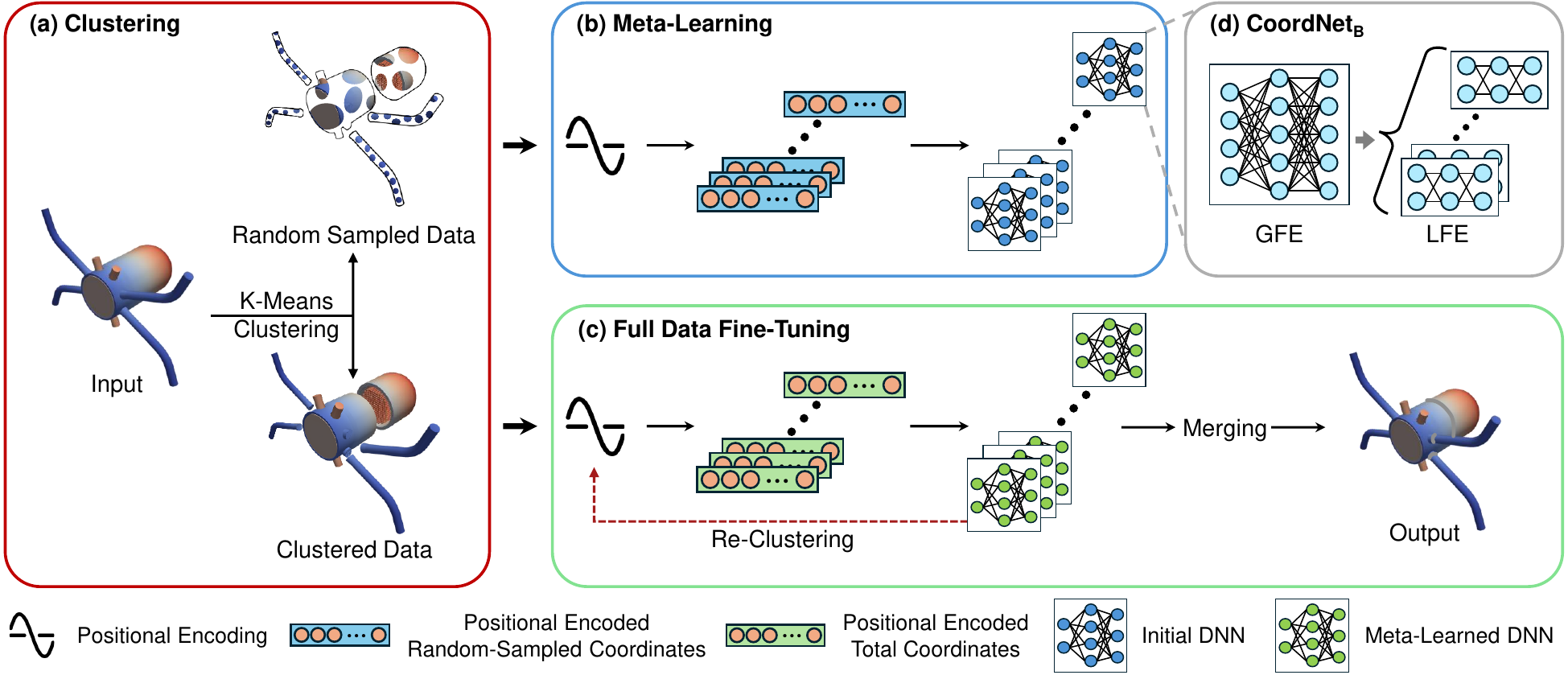}
    \caption{
    The 
    overview
    of MC-INR: (a) The input data is spatially partitioned using k-means clustering, and random sampling is performed within each cluster for meta-learning. (b) The sampled data are passed through positional encoding and used to train an initial network via meta-learning. (c) The meta-learned model is then fine-tuned using the full data of each cluster. Re-clustering is performed if the residual exceeds a threshold. (d) The proposed CoordNet$_B$ is the backbone DNN model, consisting of global and local structure feature extractors (GFE and LFE). Finally, the outputs from all clusters are merged to produce the final encoding result.
    }
    \label{fig:overall}
\end{figure*}

\subsection{INR for Scientific Simulation Data}
\label{sec:rel_INR}

Traditionally, spatial data structures, such as rectilinear grids and polygonal meshes, have been used for scientific data visualization~\cite{muigg2007scalable, shen2006visualization,wang2018visualization,yu2010situ}. 
%
%
%
%
%
%
Recently, numerous studies have emerged using INR~\cite{sitzmann2020implicit}, representing data as continuous functions.
A pioneering work by Han and Wang~\cite{han2022coordnet} proposed coordinate-based data encoding using a neural network, \textit{CoordNet}, improving the model's representation performance on 
time-varying data.
Wurster \etal~\cite{wurster2023adaptively} suggested adaptively placed feature grids in the encoder to effectively capture spatial characteristics.
Tang and Wang~\cite{tang2024stsr} and Jiao \etal~\cite{jiao2024ffeinr} introduced the INR methods based on deep neural networks (DNN) to handle the spatial and temporal variables.
%
Wang \etal~\cite{wang2024neural} proposed a hash encoder~\cite{muller2022instant}-based method combined with finite differences to capture global and local information and improve boundary conditions.
%
%
While prior works mainly focus on single-variable data in structured grids, limiting fine structure representation, we address this by exploring multivariate data on unstructured grids using a branched layer.

%% file: section/3_method.tex


\label{sec:method}

MC-INR applies k-means clustering to partition the data and performs meta-learning using sampled points from each cluster. 
Each cluster is then trained using all its data points, guided by the acquired meta-knowledge.
To handle multivariate data, MC-INR employs a branched network layer. 
The 
overview
of MC-INR is shown in~\cref{fig:overall}, and the detailed methodology is described below.

\subsection{Meta Learning and Clustering}
\label{sec:method_meta_clu}

Directly training on large datasets can make it difficult to effectively encode the data, as the underlying characteristics may be complex and hard to capture.
To address this, we adopt the k-means clustering strategy inspired by Liu \etal~\cite{liu2024uginr}.
This method partitions the dataset, spatially coherent sub-regions based on spatial coordinates, preserving local structures and enabling the model to capture localized temporal variations.
The cluster partition is defined as:
\begin{equation}
    C = \arg\min_{C} \sum_{i=1}^{K} \sum_{\mathbf{x} \in C_i} || \mathbf{x} - \boldsymbol{\mu}_i ||^2,
\end{equation}
where $C=\{ C_1, C_2, \dots , C_K \}$ represents the set of $K$ clusters, $\mathbf{x}$ denotes a data point which is spatial coordinate, and $\boldsymbol{\mu}_i$ is the centroid of cluster $C_i$.
The number of clusters, denoted as $K$, is set to 20.
Each cluster is assigned an independent DNN and process, enabling concurrent execution across multiple GPUs for efficient resource utilization and improved training throughput.

To enhance the model's performance, we further introduce a residual-based dynamic re-clustering mechanism.
After training the initial cluster, we compute the residual by the mean squared error (MSE).
If the residual of the cluster exceeds a threshold $\tau = $ 5e-4, the cluster is subdivided into two sub-clusters to preserve local temporal variations.
Otherwise, the training process is terminated.
%
%
%
%
%
Each sub-cluster inherits the model parameters from the cluster before division, preserving learned knowledge while refining representations in high-error regions.
Note that each cluster is trained by an independent DNN.
With this strategy, our model effectively captures localized temporal variations.
%


INRs typically struggle to represent 
complex structures accurately
; to mitigate this, we employ meta-learning~\cite{finn2017model}, aiming to adapt to the complexities of scientific data.
%
Inspired by Meta-INR~\cite{yang2025meta}, which performs sub-sampling by selecting points at regular intervals along each axis in structured grids, we extend this idea to unstructured grids.
However, since unstructured data lacks regularity, interval-based sampling is not feasible. 
Therefore, we adopt a random sampling strategy, selecting points across the domain without relying on fixed intervals.
This approach allows us to efficiently construct representative sub-sampled datasets suitable for meta-learning in unstructured grids.
These meta-learning and clustering approaches are illustrated in ~\cref{fig:overall} (a), (b), and (c).

\subsection{Branched CoordNet}
\label{sec:method_branched}

Scientific simulation data typically includes multiple physical variables; however, most existing methods are designed to handle only a single variable, limiting to represent multivariate datasets effectively.
%
To address this issue, 
we propose a branched layer network based on CoordNet~\cite{han2022coordnet}.
It consists of SIREN~\cite{sitzmann2020implicit} network with residual blocks, which has deeper network depth for accurate encoding.
%
Our network architecture is an extension of CoordNet, referred to as Branched CoordNet (CoordNet$_B$), consisting of 
%
global and local structure feature extractors (GFE and LFE).
The GFE captures global structural features, such as overall patterns, while the LFE is designed to learn local features specific to each physical variable, including fine-grained details.
Furthermore, to facilitate stable training and enhance the learning of fine-grained details, we leverage sine activation functions~\cite{sitzmann2020implicit} along with skip connections within the network architecture.
It is illustrated in~\cref{fig:overall} (d) and is formulated as follows:
\begin{equation}
    \begin{aligned}
    \mathbf{h}_{0} &= \mathbf{PE}([\,x, y, z, t\,]), \\
    \mathbf{y}_j &= \mathbf{LFE}_{j}(\mathbf{GFE}(\mathbf{h}_{0})), \;\; j \in \{1, 2, \cdots, M  \}, \\
    \end{aligned}
\end{equation}
where $\mathbf{h}_{0}$ and $\mathbf{PE}$ denote the input and positional encoding, respectively.
$x, y, z, t$ refers to the coordinate at time step $t$.
$\mathbf{GFE}$ represents the GFE and $\mathbf{LFE}_j$ denotes the $j$-th LFE, where $M$ is the total number of variables (e.g., temperature, pressure, etc), and $\mathbf{y}_j$ is the $j$-th output variable from the network.
%
%
The feature extractor is formulated as follows:
\begin{equation}
    \mathbf{FE}_{l} = \psi(\mathbf{W}_{l}\,\mathbf{h}_{l-1} + \mathbf{b}_{l}), \;\; l \in \{1, 2, \cdots, N \},
\end{equation}
where $\mathbf{FE}_{l}$ represents the $l$-th feature extractor layer consisting of the $\mathbf{GFE}$ and $\mathbf{LFE}_{j}$, where $N$ is number of layers.
$\mathbf{h}_{l-1}, \psi, \mathbf{W}_l$ and $\mathbf{b}_l$ correspond to feature map, sine activation function, weight, and bias at the $l$-th layer, respectively.
%
%
%
%
Training is conducted independently for each cluster $C_i$, and to optimize each cluster, we employ MSE as a loss function.
By leveraging branched layers within the network, this approach further facilitates efficient joint learning of multiple variables within a single model.


%% file: section/4_exp.tex
\subsection{Implementation Details}

\noindent\textbf{Datasets}. We used a variety of multivariate scientific simulation datasets, including in-house datasets, ROCOM, CUPID, and synthetic datasets provided by our collaborator.
%
%
Each dataset is defined on an unstructured tetrahedron mesh.  
%
%
The dataset information, used in our experiments, is presented in~\cref{tab:dataset}.
\noindent\textbf{Implementation}. We conducted experiments using PyTorch 2.1.0 on four NVIDIA RTX A6000 GPUs.
The meta-learning training hyperparameters are set to be the same as those used in Meta-INR~\cite{yang2025meta}.
%
The network consists of a total of 11 layers, comprising 5 layers in the GFE and 6 layers in the LFE, where each layer is implemented using the residual block design introduced in CoordNet~\cite{han2022coordnet}. 
%
The width of the network was adjusted to the size and characteristics of each dataset.
%
We performed the main training until convergence using the Adam optimizer with an initial learning rate of 5e-5.
The convergence condition is defined as no decrease in training loss over 30 epochs.
A learning rate scheduler is applied, reducing the learning rate by a factor of 0.92 every 30 epochs.
Note that we used 30\% of the total points that were randomly sampled in each epoch during the main training.
We used peak signal-to-noise ratio (PSNR), normalized root mean squared error (NRMSE), and coefficient determination (R$^2$) for quantitative evaluation.
%
For the quantitative comparison, we average the result values across all variables and timesteps.
The bold text indicates the best result. 

\input{figures/time_comparison/time_comaprison_results}

\subsection{Results}

\noindent\textbf{Quantitative results}. \cref{tab:results} presents the experimental results comparing with previous methods across various datasets.
%
We selected baseline methods that are independent of grid structure and can be adapted to support multivariate outputs, ensuring a fair comparison with our approach.
%
MC-INR consistently outperforms prior methods, demonstrating superior accuracy.
Specifically, without meta-learning and clustering, our network architecture, CoordNet$_B$, achieves better performance than SIREN~\cite{sitzmann2020implicit} and CoordNet across all three metrics, highlighting the effectiveness of the branched layer for multivariate data.
By integrating meta-learning and clustering, MC-INR further improves performance, achieving superior results across all metrics.
This improvement stems from its ability to efficiently capture local patterns and complex structures.
%
%
We observed that SIREN and CoordNet yield negative R$^2$ scores on the Synthetic dataset, due to an outlier in one of the output channels.
We suspect this is a limitation of multichannel vector regression using DNN, though further investigation is needed.
%
Although our method requires the largest model weight size, we observed that baseline models with similar sizes still show lower performance compared to ours, making our approach the only viable option for reducing encoding error.
%
%
%
%
%
%
%
To further demonstrate training efficiency, we present the PSNR results over training time in~\cref{fig:time_results}.
Within the same training duration, MC-INR has achieved a higher PSNR than other methods,  
which indicates that
MC-INR is effective and efficient for multivariate data encoding.

\input{tables/table_dataset}

\input{tables/table_results}

\input{figures/Qualitative_1/qualitative1_resultsv3}

\input{tables/table_compression}
\noindent\textbf{Compression results}. To demonstrate the compression performance of MC-INR's, we compare PSNR at different compression ratios \HS{(CR, raw data size / model's weight size) in~\cref{tab:compression_results}}.
%
%
We compared our method to SZ3~\cite{liang2022sz3}, 
commonly used for scientific data compression that achieves high CR while preserving user-defined error bounds.
For a fair comparison, we evaluate SZ3 and MC-INR under two aligned conditions: one with similar CRs to assess differences in PSNRs, and the other with similar PSNRs to assess differences in CRs.
%
%
As shown in~\cref{tab:compression_results}, MC-INR achieved better performance in both cases, yielding higher PSNR at similar CRs and higher CR at similar PSNRs.
%
%
%
%



\noindent\textbf{Qualitative results}. Visual comparison with previous methods is illustrated in~\cref{fig:qul_results1}.
The ground truth (GT) rendered multivariate datasets by a tetrahedral mesh, and other methods visualized the error map
%
illustrating the absolute difference between the GT and the prediction, with lower errors shown in blue and higher errors in red, highlighting regions of inaccuracy.
%
%
SIREN and CoordNet exhibit high errors on the synthetic data, while CoordNet$_B$ exhibits significantly lower errors compared to them. 
MC-INR further reduces the error of CoordNet$_B$, demonstrating its effectiveness in encoding 
multivariate datasets.
%

\input{tables/table_recluster}
\noindent\textbf{Ablation study}. We evaluate the effectiveness of the residual-based dynamic re-clustering mechanism on the ROCOM dataset (\cref{tab:recluster_results}).
When re-clustering is applied, the model achieves a 2.59 dB improvement in PSNR compared to without re-clustering, along with slight gains in the other two metrics.
This improvement indicates that re-clustering enables the model to enhance the capture of specific local structures.
Therefore, the re-clustering mechanism is effective in improving the model's representational ability.

\subsection{Limitations}
Due to the use of multiple networks (one per cluster), our method incurs higher memory consumption compared to approaches using a single network. 
Additionally, non-overlapping clusters may introduce boundary artifacts, which can become noticeable when zooming into local regions near cluster edges. Introducing simple boundary overlaps could mitigate this issue. 
Despite these limitations, our method achieves superior encoding performance compared to previous works.

%% file: figures/time_comparison/time_comaprison_results.tex
\begin{figure*}[t]
    \centering
    \renewcommand{\arraystretch}{0.1}
        \resizebox{0.99\linewidth}{!}{
        \begin{tabular}{c@{\hskip 0.005\linewidth}c@{\hskip 0.005\linewidth}c@{}}
        
            \includegraphics[width=0.33\linewidth]{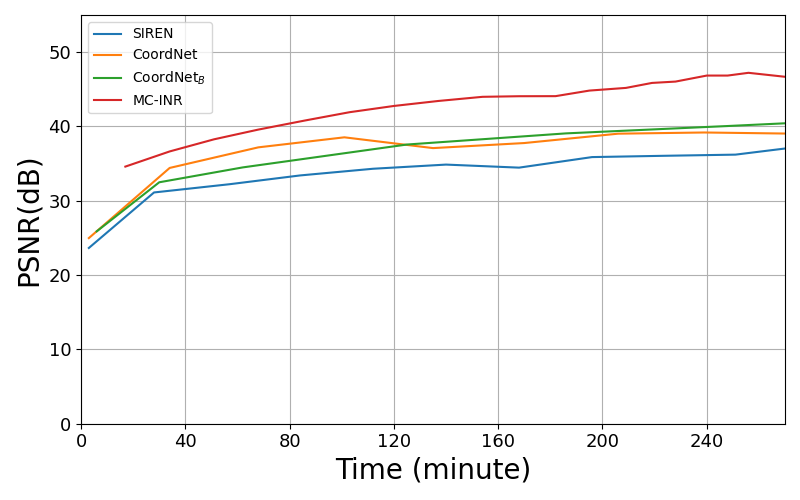} & 
            \includegraphics[width=0.33\linewidth]{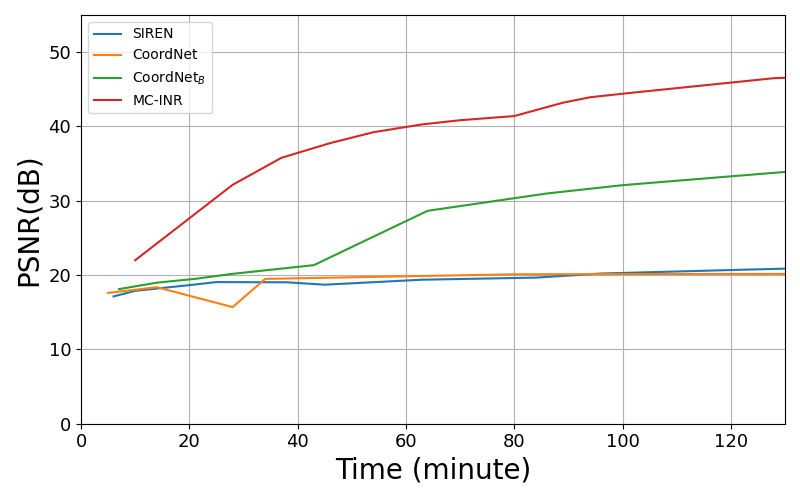} & \includegraphics[width=0.33\linewidth]{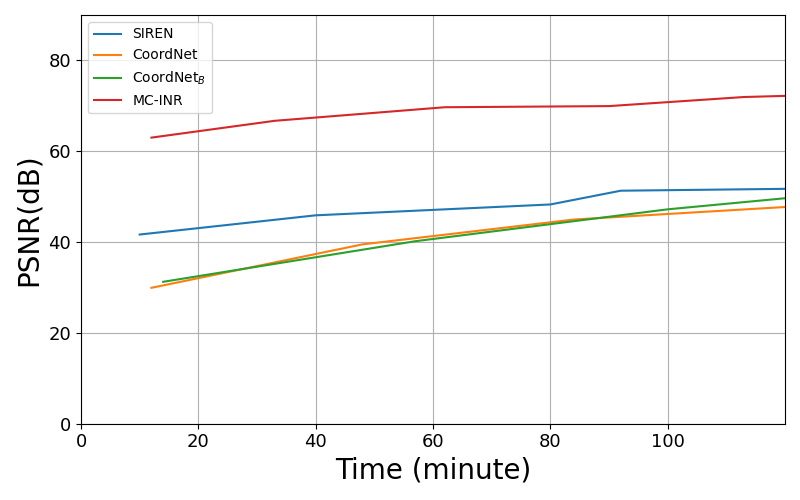} \\

            ROCOM &
            CUPID & 
            Synthetic \\

        \end{tabular}
        }
        \caption{
            Comparison of encoding performance over time. PSNR is recorded per epoch, with curves starting at different times due to varying training durations. Each method is trained until convergence.
        }
        \label{fig:time_results}
\end{figure*}

%% file: tables/table_dataset.tex
\begin{table}
\caption{
Experimental datasets overview. $\rho$ and $t$ denote the number of spatial coordinates per timestep and the number of timesteps, respectively. 
}
\label{tab:dataset}
\centering
\resizebox{0.70\linewidth}{!}{
\begin{tabular}{l|cc}
\Xhline{2\arrayrulewidth}
Dataset & \# of points ($\rho$ $\times$ $t$) & \# of variables\\ \hline \hline
ROCOM                          & 12,440,366 $\times$ 5     & 8\\  
CUPID                          & 1,279,375  $\times$ 30    & 11\\
Synthetic                      & 5,253,750  $\times$ 30    & 4\\
\Xhline{2\arrayrulewidth}
\end{tabular}
}
\end{table}

%% file: tables/table_results.tex
\begin{table}
\caption{
Quantitative comparison of data encoding results. \HS{Size indicates the model's weight, and for our method, it refers to the total size summed over all clusters.} CoordNet$_B$ means our branched network, which is without meta-learning and clustering.
}
\label{tab:results}
\centering
\resizebox{0.99\linewidth}{!}{
\begin{tabular}{ll|ccc|c}
\Xhline{2\arrayrulewidth}
Dataset & Method & PSNR$\uparrow$ & NRMSE$\downarrow$ & R$^2$$\uparrow$ & \HS{Size (KB)}\\ \hline \hline
\multirow{4}{*}{ROCOM}            & SIREN~\cite{sitzmann2020implicit}       & 38.59         & 0.0161                                        & 0.8756      & \HS{530}\\
                                  & CoordNet~\cite{han2022coordnet}         & 39.51         & 0.0129     & 0.8133       & \HS{23215}\\
                                  & CoordNet$_B$          & 47.65           & 0.0073        & 0.9765     & \HS{10372}\\
                                  & MC-INR                & \textbf{54.10}  & \textbf{0.0042}            & \textbf{0.9923}      & \HS{33600}\\ \hline           
\multirow{4}{*}{CUPID}            & SIREN                 & 21.63           & 0.1656          & 0.5659                                      & \HS{304}\\
                                  & CoordNet              & 20.51           & 0.1732          & 0.4733    & \HS{13095}\\
                                  & CoordNet$_B$          & 36.99           & 0.0160          & 0.8294    & \HS{6231}\\
                                  & MC-INR                & \textbf{51.55}  & \textbf{0.0033} & \textbf{0.9975}     & \HS{27240}\\ \hline
\multirow{4}{*}{Synthetic}        & SIREN                 & 52.04           & 0.0039          & -0.3349                                     & \HS{528}\\
                                  & CoordNet              & 50.39           & 0.0038        & -8.4193     & \HS{23199}\\
                                  & CoordNet$_B$          & 62.16           & 0.0010          & 0.7958    & \HS{9627}\\
                                  & MC-INR                & \textbf{76.54} & \textbf{0.0002} & \textbf{0.9827}          & \HS{30180}\\ 
\Xhline{2\arrayrulewidth}
\end{tabular}
}
\end{table}

%% file: figures/Qualitative_1/qualitative1_resultsv3.tex
\begin{figure*}[t]
    \centering
    \renewcommand{\arraystretch}{0.1}
        \resizebox{0.95\linewidth}{!}{
        \begin{tabular}{@{}c@{\hskip 0.04\linewidth}c@{\hskip 0.005\linewidth}c@{\hskip 0.005\linewidth}c@{\hskip 0.005\linewidth}c@{\hskip 0.005\linewidth}c@{\hskip 0.005\linewidth}c@{}}

            \rotatebox{90}{\parbox{2.4cm}{\centering ROCOM}} &
            \includegraphics[width=0.19\linewidth]{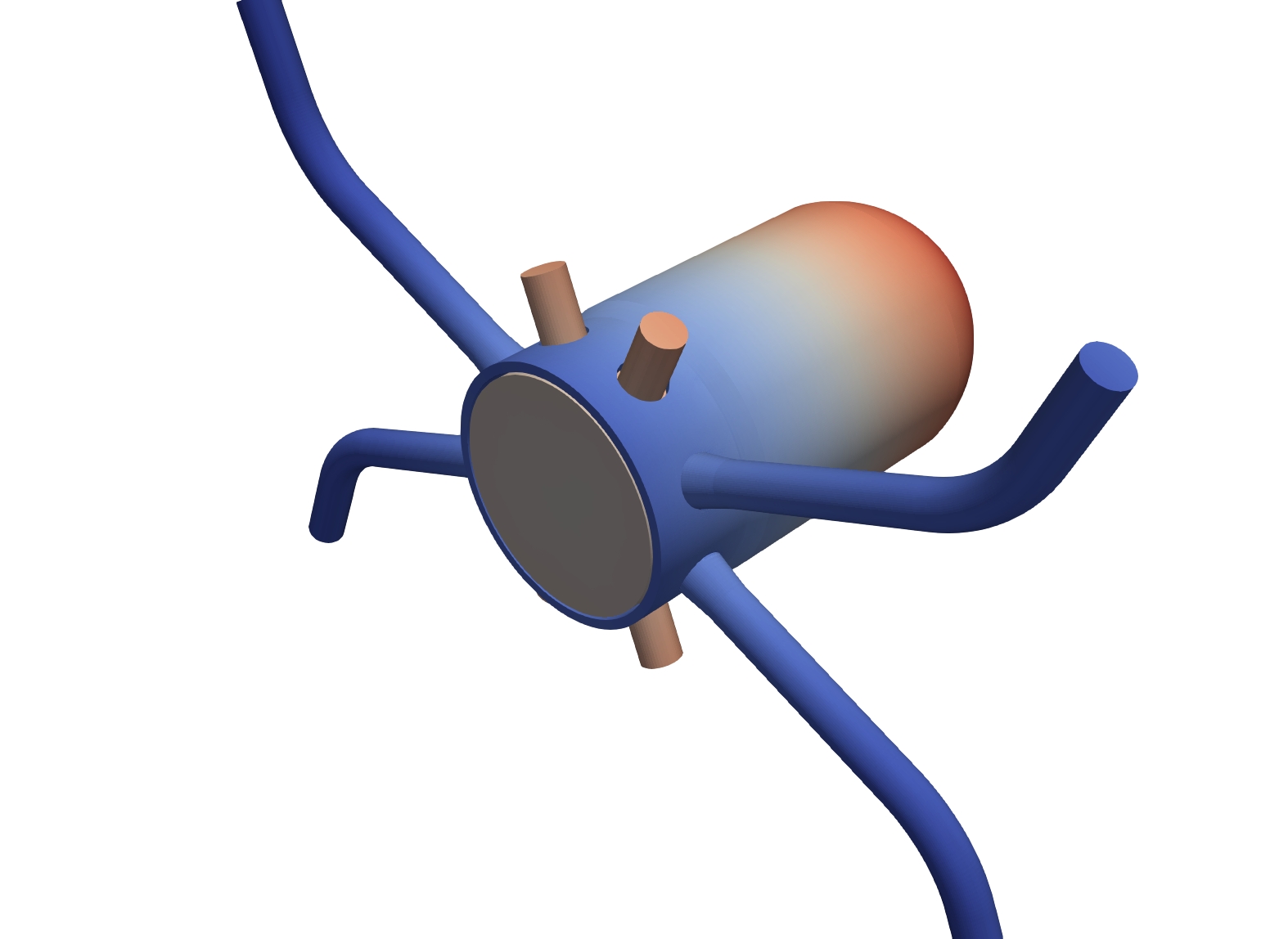} & 
            \includegraphics[width=0.19\linewidth]{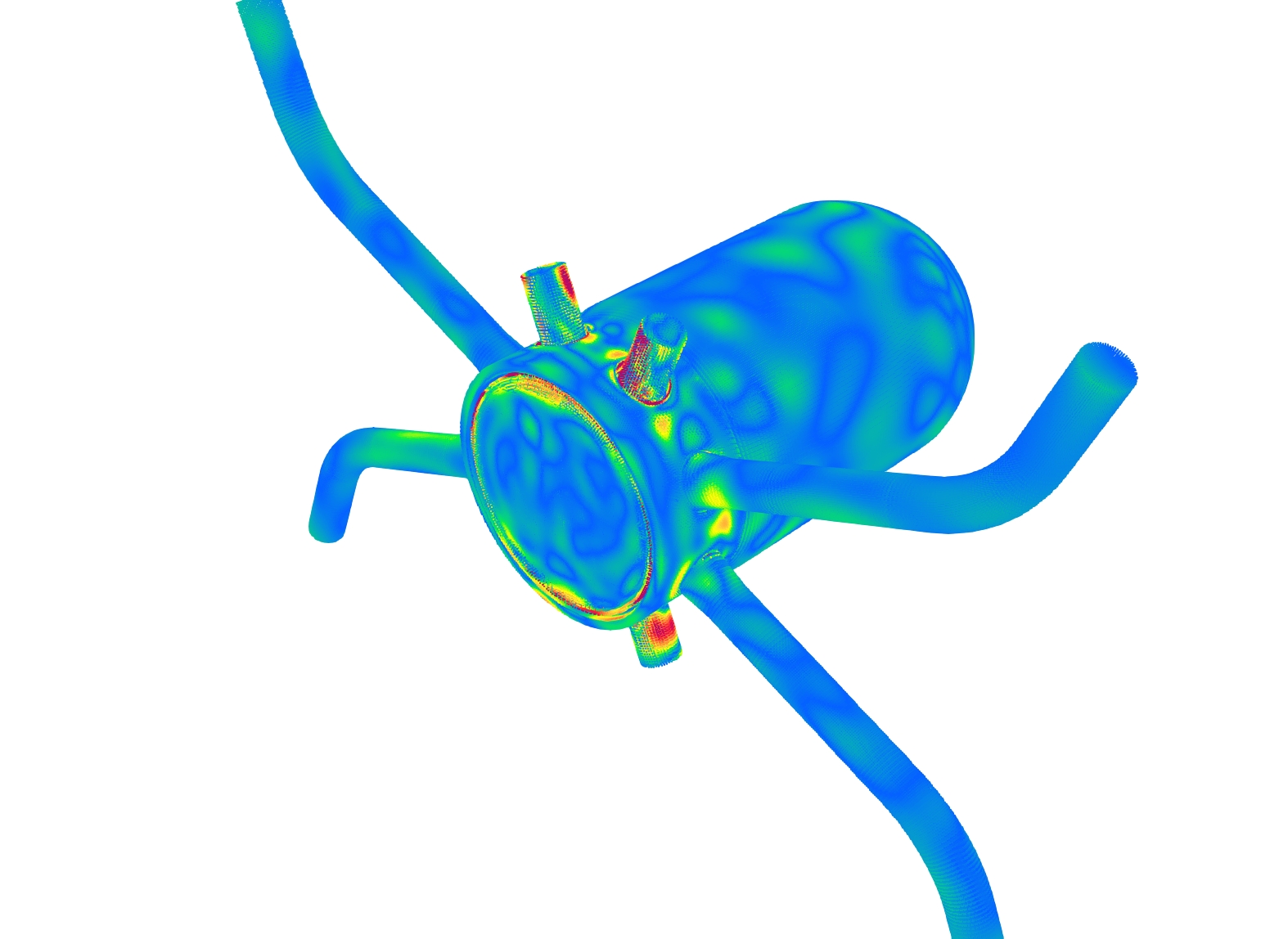} &
            \includegraphics[width=0.19\linewidth]{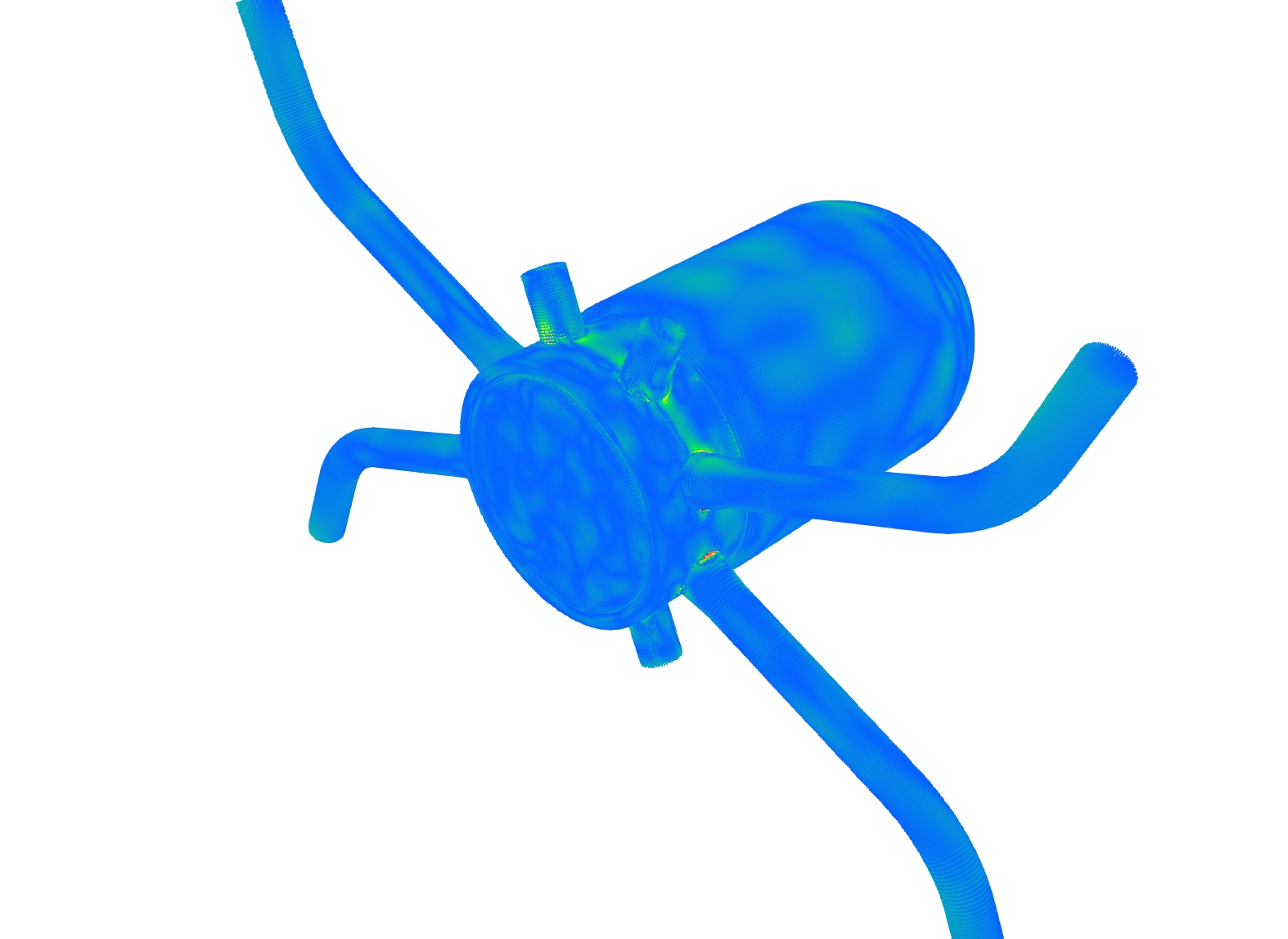} &
            \includegraphics[width=0.19\linewidth]{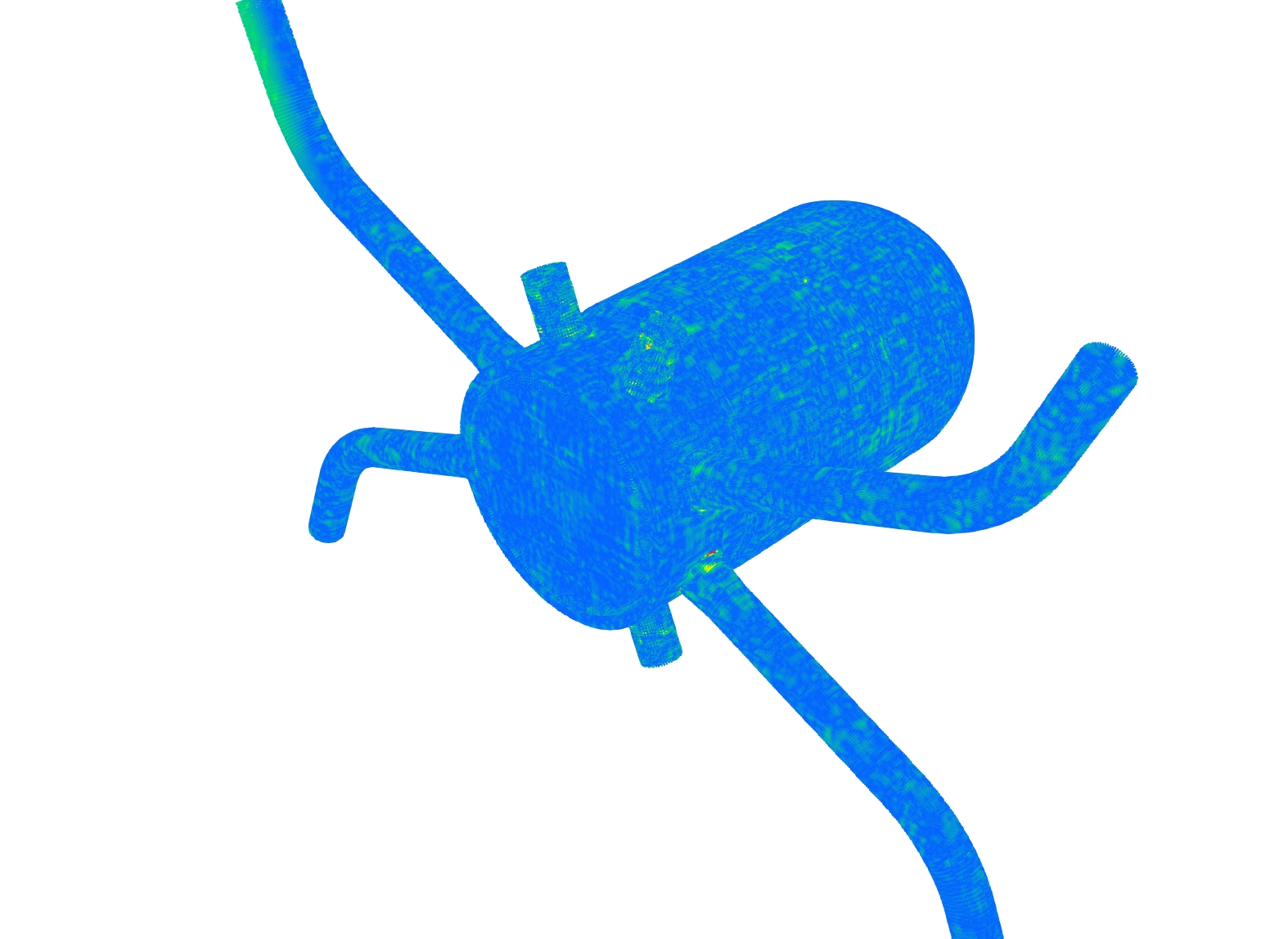} &
            \includegraphics[width=0.19\linewidth]{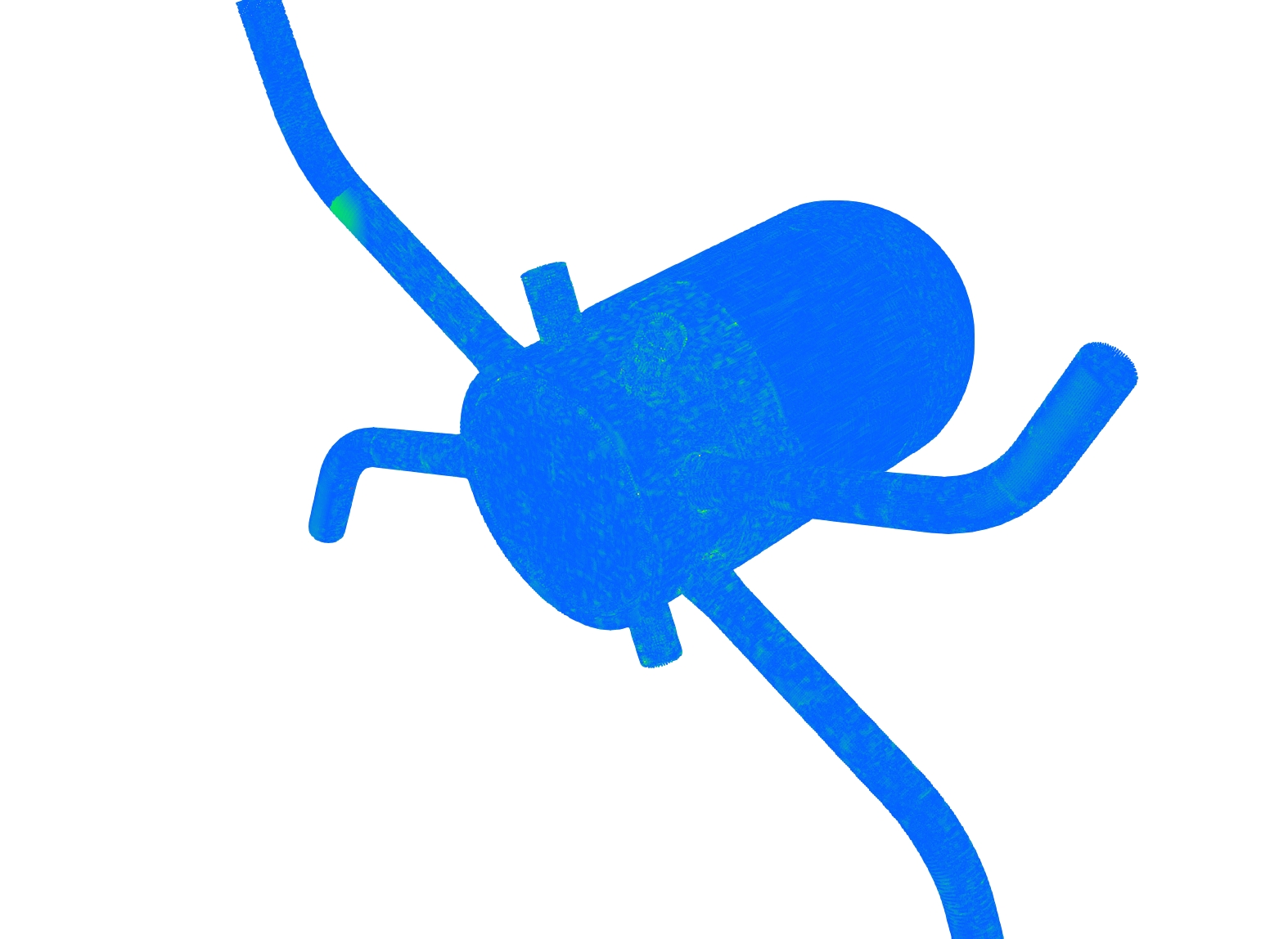} &
            \includegraphics[width=0.04\linewidth]{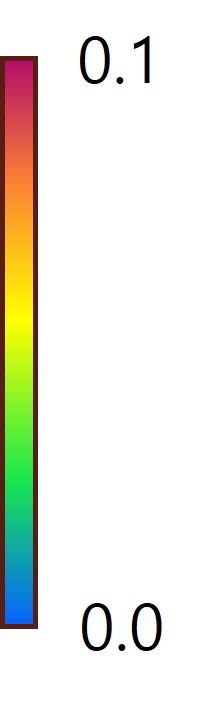} \\

            \rotatebox{90}{\parbox{2.6cm}{\centering CUPID}} &
            \includegraphics[width=0.19\linewidth]{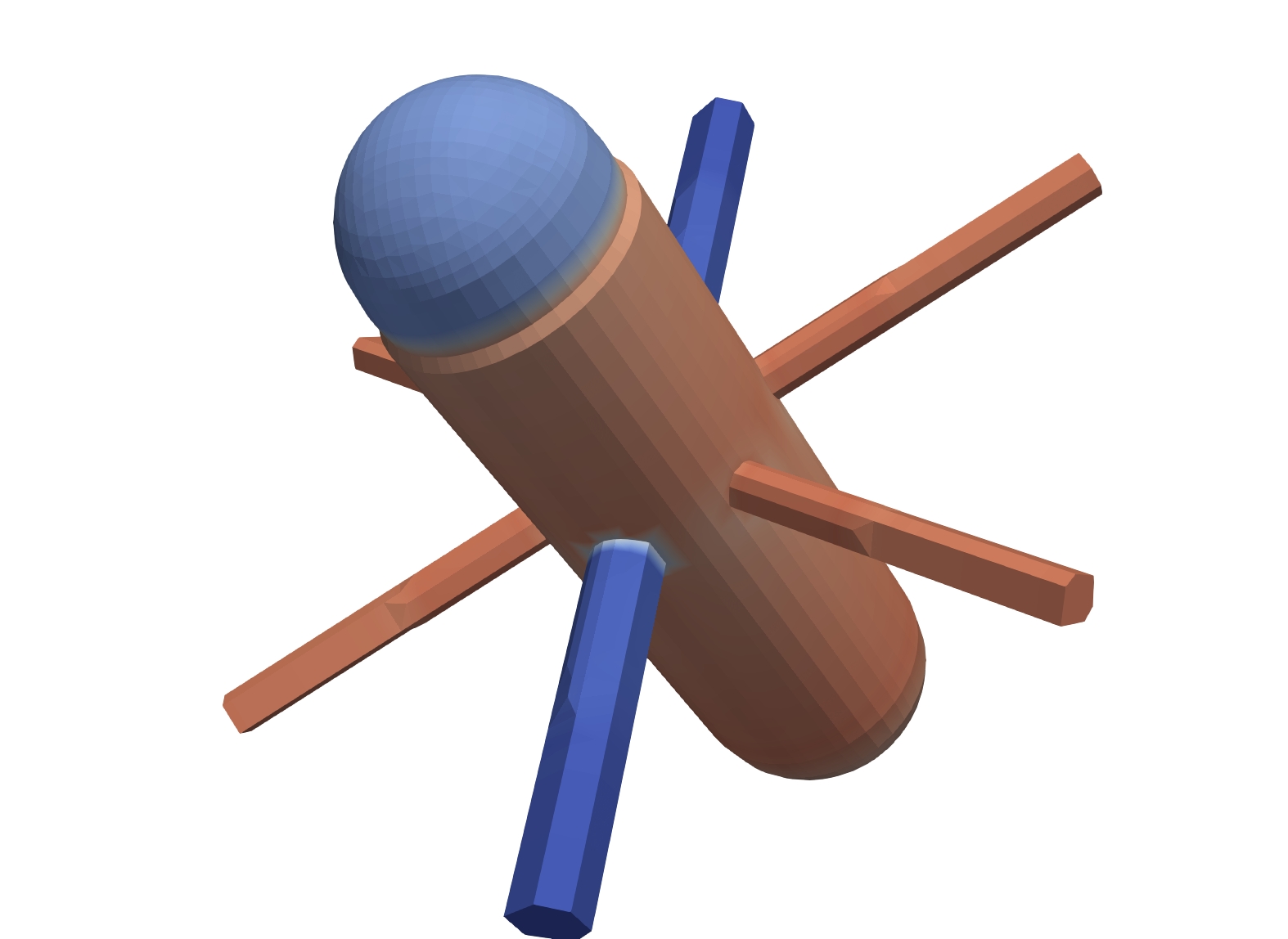} &
            \includegraphics[width=0.19\linewidth]{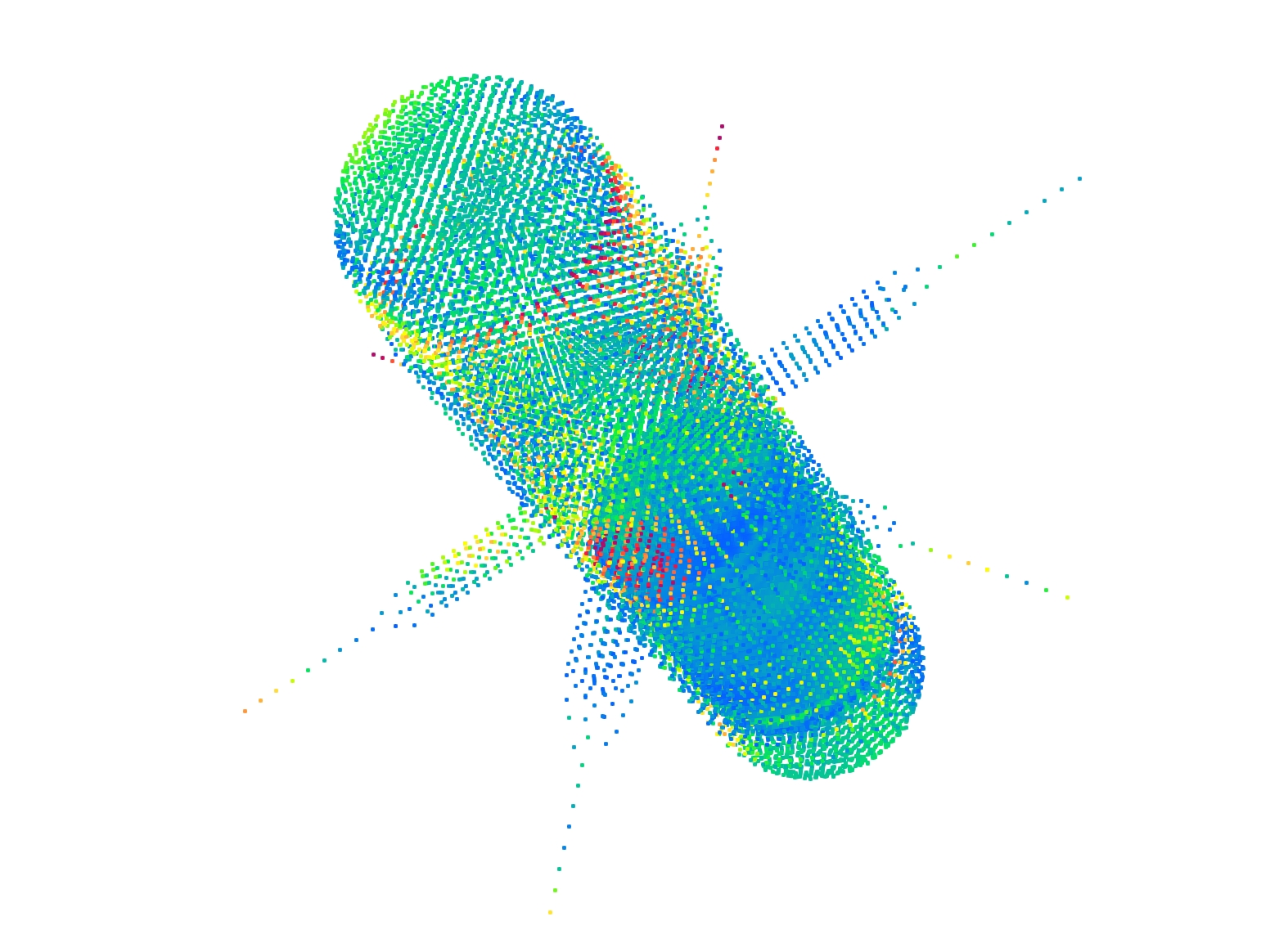} &
            \includegraphics[width=0.19\linewidth]{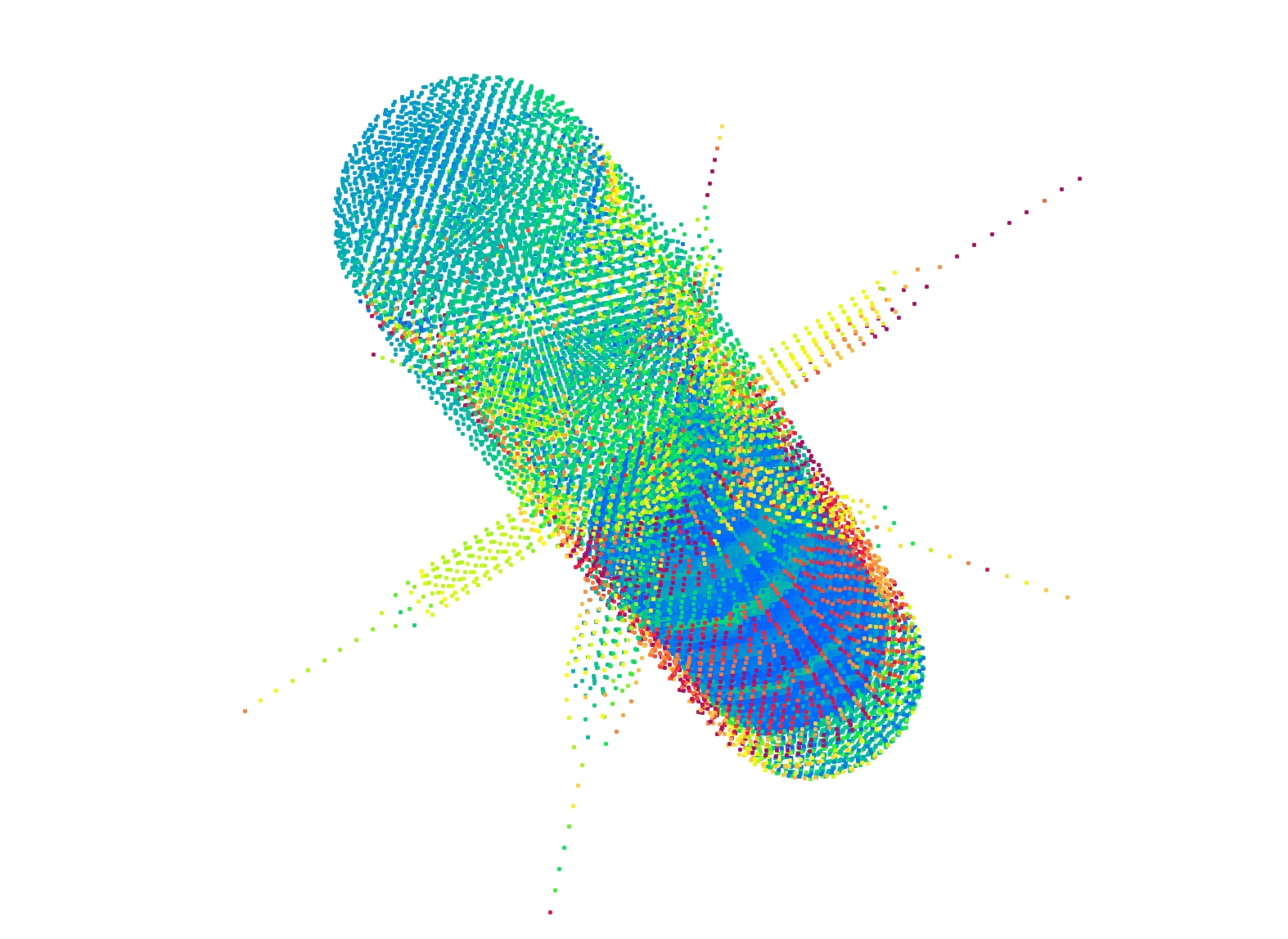} &
            \includegraphics[width=0.19\linewidth]{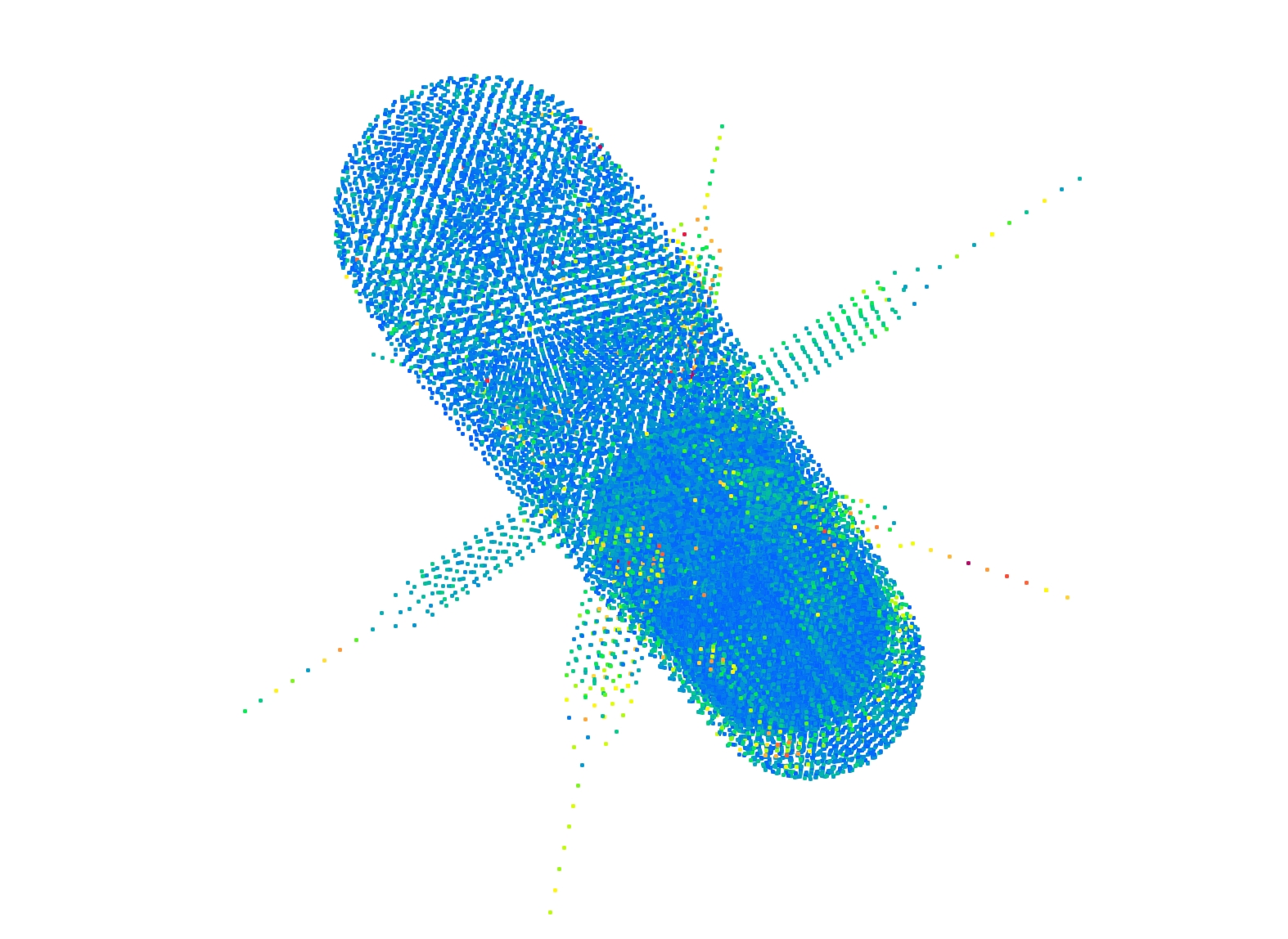} &
            \includegraphics[width=0.19\linewidth]{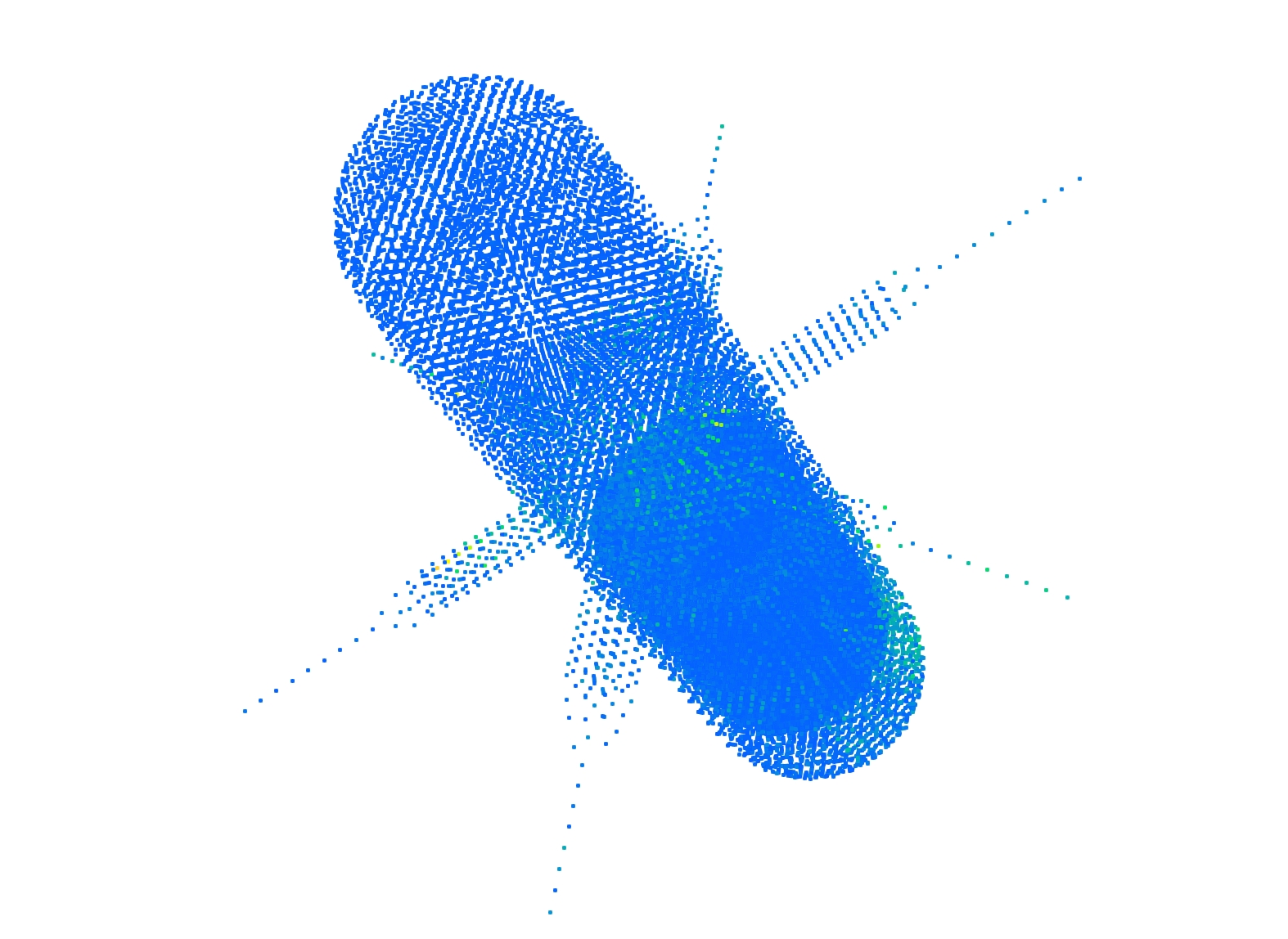} &            \includegraphics[width=0.04\linewidth]{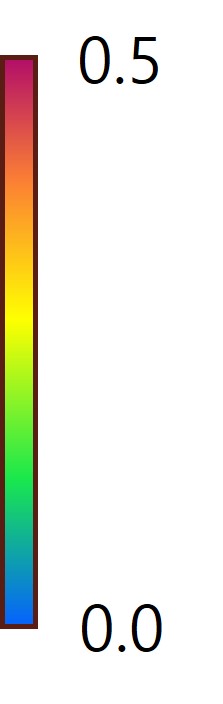} \\

            \rotatebox{90}{\parbox{1.9cm}{\centering Synthetic}} &
            \includegraphics[width=0.19\linewidth]{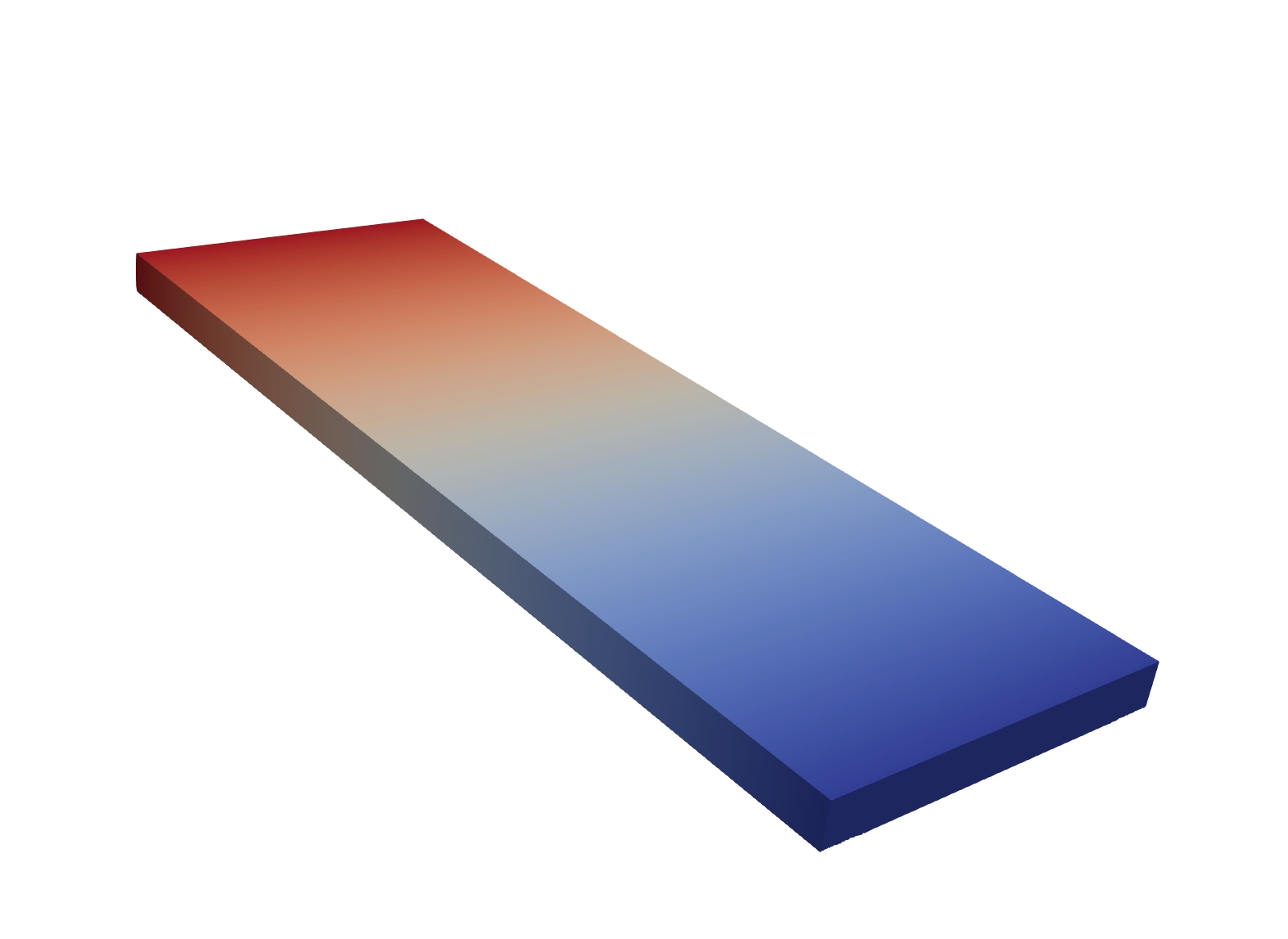} &
            \includegraphics[width=0.19\linewidth]{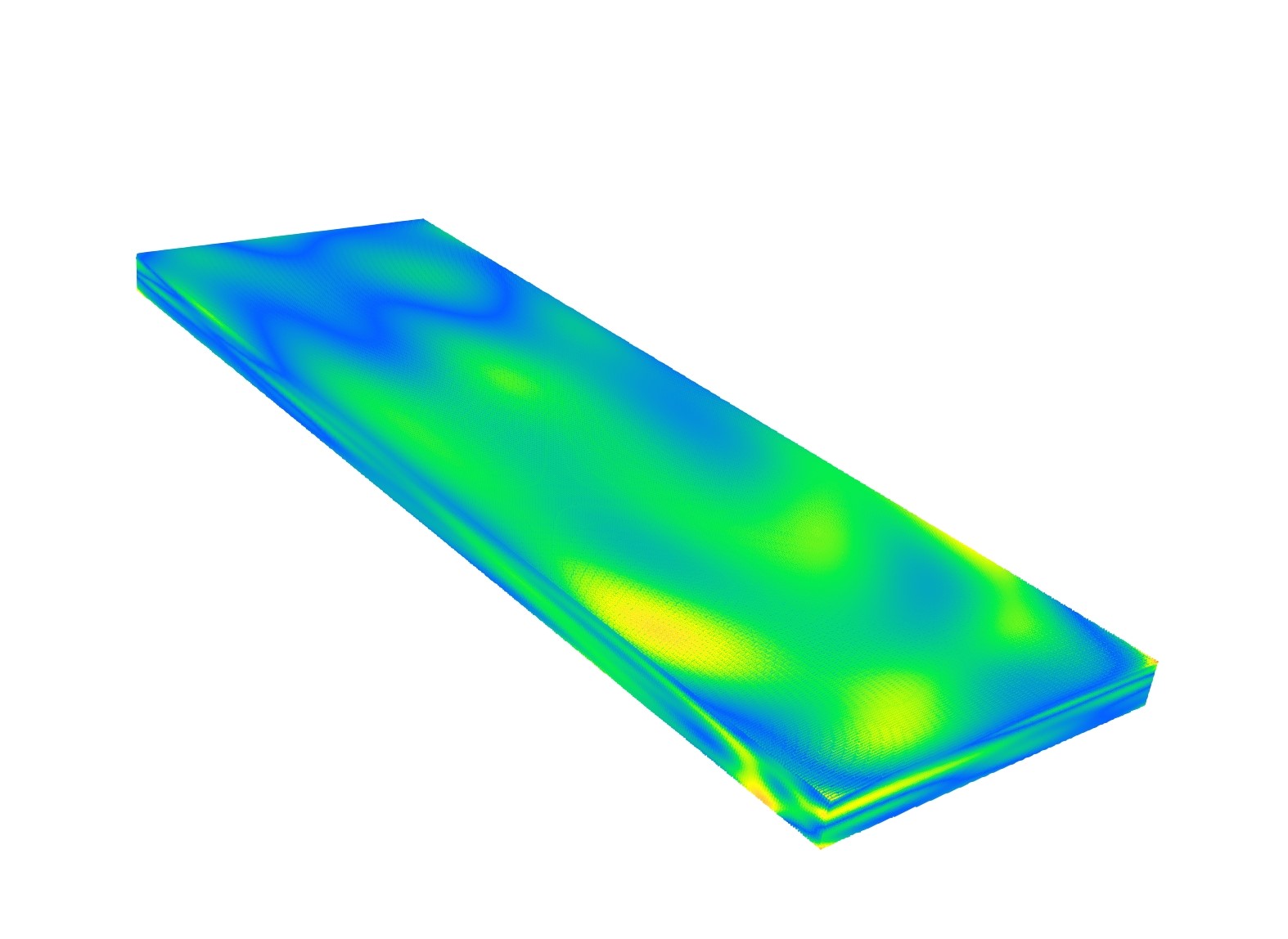} &
            \includegraphics[width=0.19\linewidth]{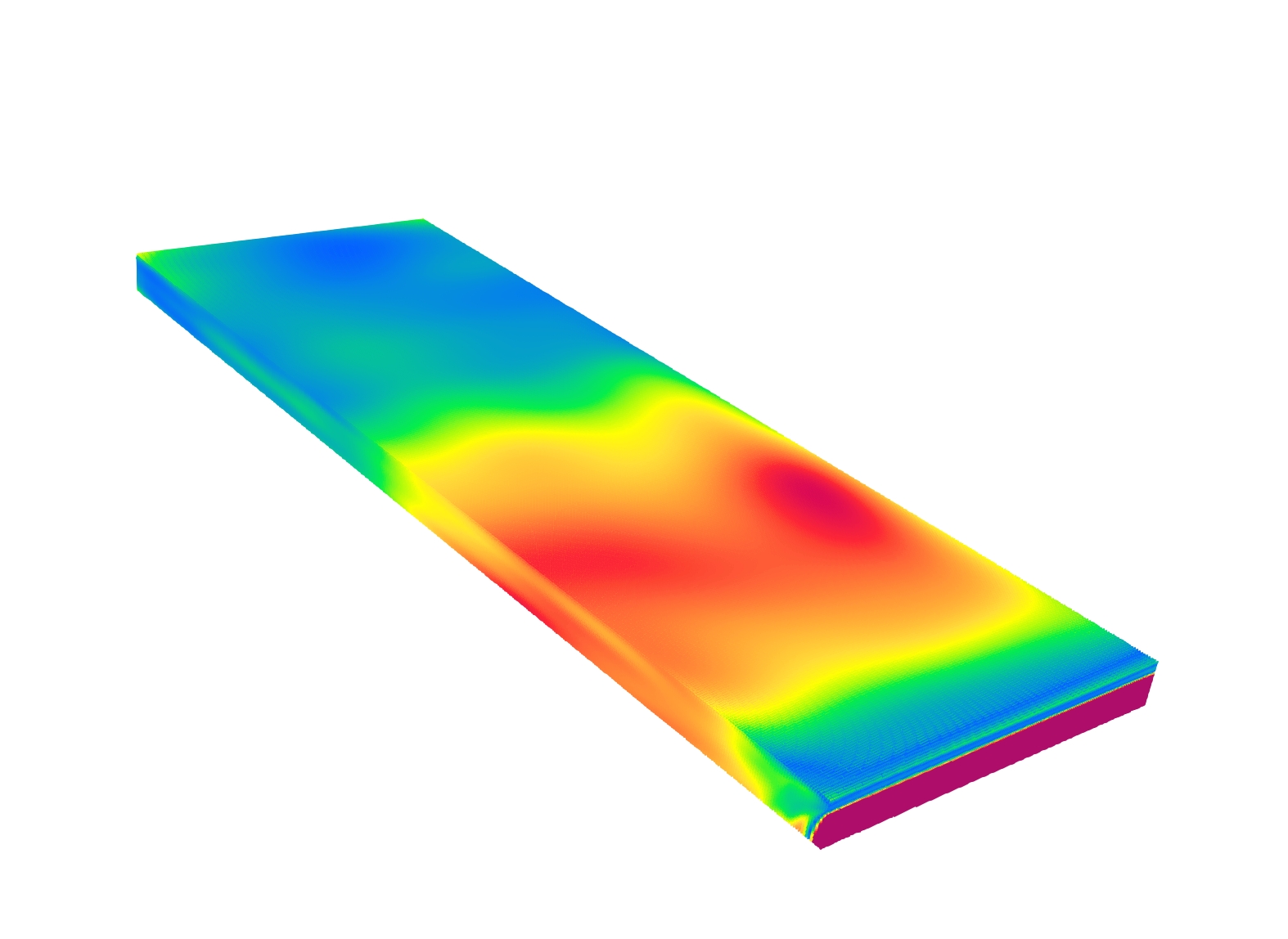} &
            \includegraphics[width=0.19\linewidth]{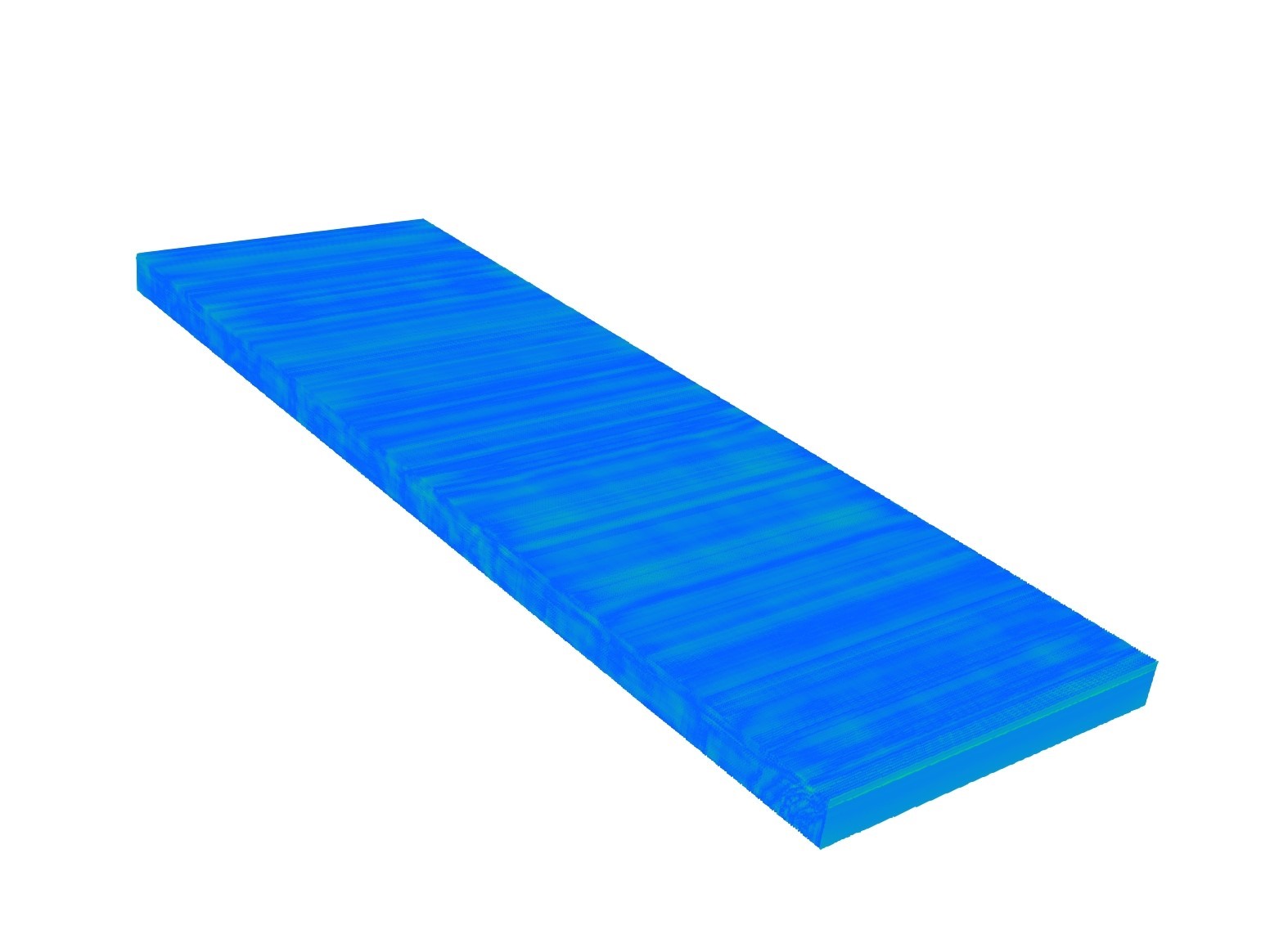} &
            \includegraphics[width=0.19\linewidth]{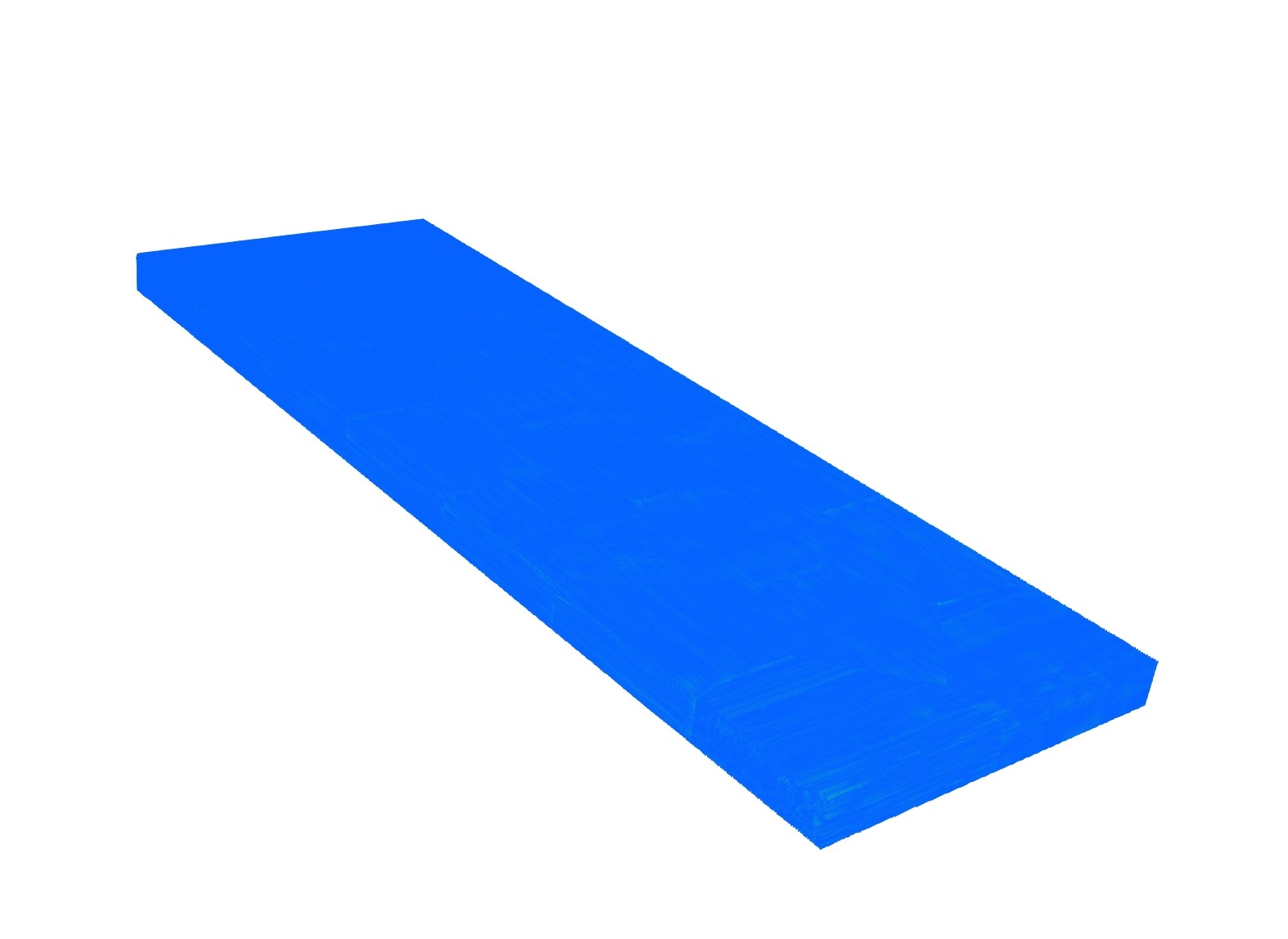} &            \includegraphics[width=0.04\linewidth]{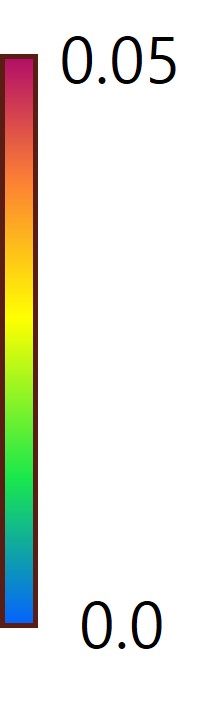} \\


            &
            GT &
            SIREN~\cite{sitzmann2020implicit} &
            CoordNet~\cite{han2022coordnet} & 
            CoordNet$_B$ &
            MC-INR \\

        \end{tabular}
        }
        \caption{Error map visualization \HS{of pressure value}. 
        Errors are computed using \HS{absolute difference},
        with lower values shown in blue and higher values in red. The ground truth (GT) refers to the reconstructed points obtained from the tetrahedral mesh.
        }
        \label{fig:qul_results1}
\end{figure*}

%% file: tables/table_compression.tex
\begin{table}
\caption{
Quantitative comparison of data compression results. We evaluated under identical PSNR and CR settings to ensure a fair comparison. CR refers to the compression ratio.
}
\label{tab:compression_results}
\centering
\resizebox{0.62\linewidth}{!}{
\begin{tabular}{ll|cc}
\Xhline{2\arrayrulewidth}
Dataset & Method & PSNR$\uparrow$   & CR$\uparrow$\\ \hline \hline                                        
\multirow{3}{*}{ROCOM}            & SZ3~\cite{liang2022sz3}        & 22.83           & 109\\
                                  & SZ3                            & 54.07           & 4.89\\
                                  & MC-INR                & \textbf{54.10}           & \textbf{110}\\ \hline
                                              
\multirow{3}{*}{CUPID}            & SZ3                   & 16.69           & 229\\
                                  & SZ3                   & 51.30           & 2.03\\
                                  & MC-INR                & \textbf{51.55}           & \textbf{230}\\ \hline

\multirow{3}{*}{Synthetic}        & SZ3                   & 36.96           & 310\\
                                  & SZ3                   & 76.05           & 14.61\\
                                  & MC-INR                & \textbf{76.54}           & \textbf{311}\\ 
\Xhline{2\arrayrulewidth}
\end{tabular}
}
\end{table}

%% file: tables/table_recluster.tex
\begin{table}
\caption{
Ablation study results for residual-based re-clustering. This experiment is conducted on the ROCOM dataset.
}
\label{tab:recluster_results}
\centering
\resizebox{0.75\linewidth}{!}{
\begin{tabular}{l|ccc}
\Xhline{2\arrayrulewidth}
Method & PSNR$\uparrow$ & NRMSE$\downarrow$ & R$^2$$\uparrow$\\ \hline \hline
                                              
w/o Re-clustering              & 51.51              & 0.0052                & 0.9886\\
w/ Re-clustering               & \textbf{54.10}     & \textbf{0.0042}       & \textbf{0.9923}\\
                                              
\Xhline{2\arrayrulewidth}
\end{tabular}
}
\end{table}

%% file: section/5_conclusion.tex


In this paper, we introduced MC-INR, a novel neural network-based framework designed to encode multivariate scientific simulation data on unstructured grids.
MC-INR leverages meta-learning and clustering strategies to effectively encode data on complex unstructured grids and capture local patterns.
The proposed residual-based dynamic re-clustering mechanism further contributes to improving the encoding accuracy.
%
%
Additionally, we introduced CoordNet$_B$, which enables joint learning of multivariate data and enhances representation quality.
As a result, MC-INR achieved superior performance in both quantitative and qualitative evaluations, demonstrating its effectiveness and robustness for scientific data visualization.
%

%
For future work, 
we plan to explore various clustering strategies that leverage both spatial structures and data distributions for higher encoding performance. 
%
%
%
Moreover, we intend to conduct in-depth validation across diverse datasets, including both structured and unstructured grid data.
%

